# **SYMOG**: learning **sy**mmetric **m**ixture **o**f **G**aussian modes for improved fixed-point quantization

Lukas Enderich[a], Fabian Timm[b], Wolfram Burgard[c]

[a] *Robert Bosch GmbH - Chassis Systems Control, Engineering Cognitive Systems*
[b] *Robert Bosch GmbH - Corporate Research, Vehicle Safety and Automated Driving*
[c] *University of Freiburg - Autonomous Intelligent Systems*

---

**Abstract**

Deep neural networks (DNNs) have been proven to outperform classical methods on several machine learning benchmarks. However, they have high computational complexity and require powerful processing units. Especially when deployed on embedded systems, model size and inference time must be significantly reduced. We propose **SYMOG** (**sy**mmetric **m**ixture **o**f **G**aussian modes), which significantly decreases the complexity of DNNs through low-bit fixed-point quantization. SYMOG is a novel soft quantization method such that the learning task and the quantization are solved simultaneously. During training the weight distribution changes from an unimodal Gaussian distribution to a symmetric mixture of Gaussians, where each mean value belongs to a particular fixed-point mode. We evaluate our approach with different architectures (LeNet5, VGG7, VGG11, DenseNet) on common benchmark data sets (MNIST, CIFAR-10, CIFAR-100) and we compare with state-of-the-art quantization approaches. We achieve excellent results and outperform 2-bit state-of-the-art performance with an error rate of only 5.71% on CIFAR-10 and 27.65% on CIFAR-100.

---

*Email address:* Lukas.Enderich@de.bosch.com (Lukas Enderich)

## 1. Introduction

Deep learning architectures with stacked non-linearities have significantly improved recent progress in machine learning. Notably, deep neural networks (DNNs) have set new minimum benchmarks in various fields of research, such as computer vision, speech recognition and object detection (Deng and Yu (2014); Lecun et al. (2015); Karki et al. (2019)). Furthermore, DNNs are now also capable from a safety perspective by providing uncertainties, as it is required for safe autonomous driving (Feng et al. (2019a,b)). However, with millions of parameters and billions of high-precision computations, DNNs require powerful and expensive processing units, which are usually unavailable on embedded devices (Krizhevsky et al. (2012); Simonyan and Zisserman (2014)).

For self-driving cars, the issue of high power consumption by DNNs becomes even more critical if the goal is both autonomous and electrified driving. Therefore, different approaches can be applied to reduce the model complexity and to make DNNs more efficient. On one hand, pruning and factorization remove single neurons or complete filters to decrease the model size (Mauch and Yang (2017, 2018)). In contrast, quantization reduces the precision (bit-size) of operands and operations, and low-bit fixed-point representations substantially reduce memory cost, inference time and energy consumption. For example, an 8-bit fixed-point multiplication requires 18.5 times less energy than a 32-bit floating-point multiplication and 4 times less computation time on Nvidia's pascal architecture (Sze et al. (2017)). DNN weights encoded with two bits (Li and Liu (2016); Zhu et al. (2016); Zhou et al. (2018)) or even one bit (Courbariaux et al. (2015); Zhou et al. (2018)) replace many multiplications by additions.

In network quantization there are two major challenges: maximizing performance and ensuring convergence. On one hand, naive post-quantization of already trained model parameters decreases test accuracy, especially with fewer than 8 bits (Lin et al. (2016)). On the other hand, gradient descent requires high-precision weights and activations to converge successfully (Guo (2018)). Training with discrete operands is an intractable problem. Nevertheless, recent approaches employ different training methods to achieve optimal performance for quantized weights.

However, many quantization functions retain high-precision scaling coefficients (Li and Liu (2016); Zhu et al. (2016); Zhou et al. (2018, 2017)), which exclude pure fixed-point arithmetic on dedicated hardware. Moreover, incre-mental retraining approaches such as weight partitioning (Zhou et al. (2018, 2017)) or multi-level training (Yin et al. (2018)), complicate the training procedure and involve additional hyperparameters.
We introduce **SYMOG** (**Sy**mmetric **M**ixture **of G**aussian modes), a gradient-based approach to train DNNs with multimodal weight distributions that enables pure fixed-point quantization with minimal quantization error and excellent performance. Initialized with a pre-trained model, our approach employs self-reliant weight adaptation by solving the learning task and the quantization constraint simultaneously during the training (see Figure 1 as motivation). The gradient procedure is easy to implement, needs comparatively few training epochs and achieves new state-of-the-art performance using ternary-valued weights (5.71% on CIFAR-10, 27.65% on CIFAR-100).

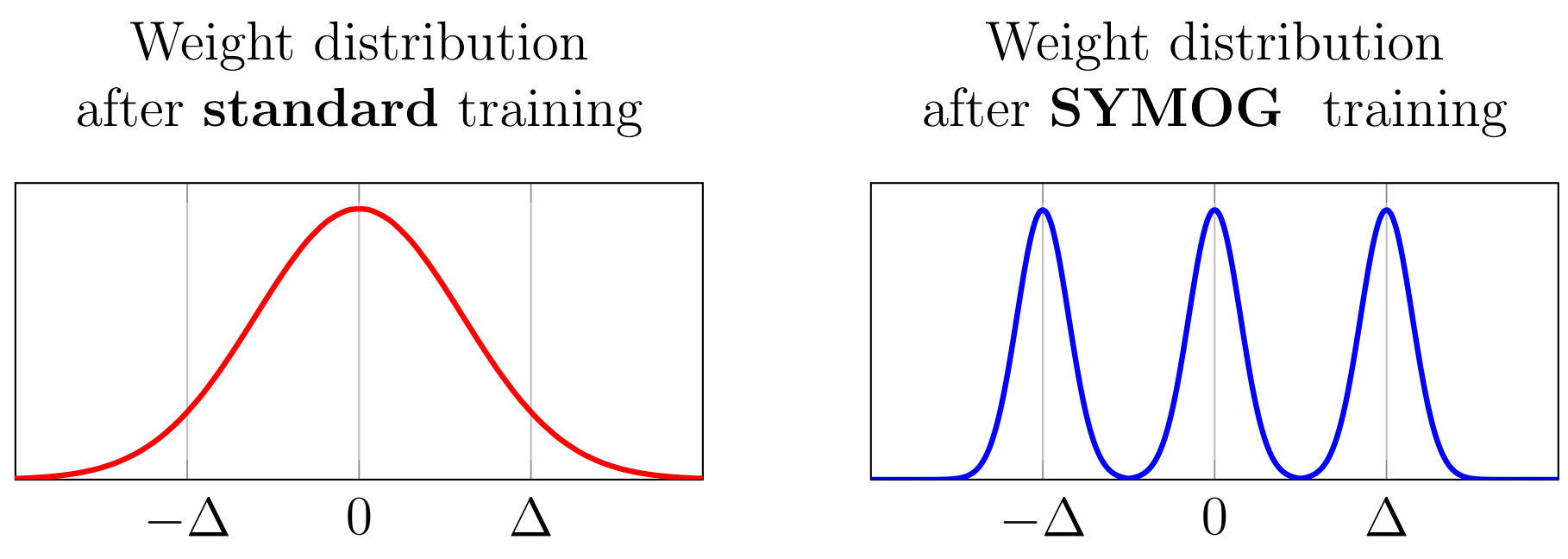

Figure 1: Simplified comparison of the weight distribution after training. With conventional training, weights are usually unimodal Gaussian distributed and, after quantization to symmetric bins, yield high quantization error and poor performance (left). In contrast, with SYMOG , each quantization bin is represented by a single Gaussian distribution such that both quantization error and error rate are minimized simultaneously (right).

## 2. Related Work

Quantization reduces the bit-size of model weights rather than their quantity, which can otherwise be obtained with pruning approaches. There are various options for quantization and these can be classified into three categories: naive quantization, hard quantization, and soft quantization (Sze et al. (2017)).

### *2.1. Naive quantization*

Naive quantization converts a pretrained floating-point model into a quantized representation by post-processing alone (Lin et al. (2016)). The training itself does not include any quantization constraint. Instead, post-processing analyzes the layer statistics to estimate a reasonable quantization function for each layer (note that after training, weights are usually unimodally Gaussian distributed, with slightly different parameters w.r.t. to the particular layer - thus, distances between these distributions and the quantization bins must be minimized). The post-processing approach is the easiest to implement but reduces accuracy tremendously, especially for low-bit quantization (Lin et al. (2016)).

### *2.2. Hard quantization*

Hard quantization methods integrate discrete weights into parts of the DNN training to reduce the loss perturbation caused by a discretized inference. For example, Courbariaux et al. (2015) use binarized weights during both forward and backward pass to compute adequate gradients, but update high-precision weights instead, which are kept during the whole optimization process. Through this, training with gradient descent converges.
Li and Liu (2016) increased the capacity of discrete models by combining ternary-valued weights and a scaling coefficient. For an optimal scaling factor, the euclidean distance between the high-precision weights and the scaled ternary weights must be approximated and solved with a threshold based ternary function. Further extensions involve two independent scaling factors to update both the continuous weights and the scaling coefficients at the same time (Zhu et al. (2016)).
Although hard quantization methods achieve convergence, they induce a gradient mismatch since different weight representations are used for the gradient computation on one hand and the update step on the other. In some cases, the accuracy is deteriorated by several percent in this way.

### *2.3. Soft quantization*

Soft quantization describes a training with real-valued weights, promoting both accurate gradient calculations and posterior distributions that are well suited for post-quantization. To achieve this, Bayesian methods have been used for model compression, e.g. by sparse posterior distributions (Ullrich et al. (2017); Louizos et al. (2017)). Recently, Achterhold et al. (2018)

introduced quantizing priors to train DNNs with multimodal weight distributions that can be quantized well afterwards. Weights with high variance are set to zero while remaining modes must lie symmetrically around zero.
Then again, a loss-error-aware quantization method for determinsistic low-bit DNNs has been proposed by Zhou et al. (2018). In this approach, the layer weights are partitioned in several steps, where one part is quantized and the remaining part is used for retraining. Furthermore, Enderich et al. (2019) introduce Gaussian priors to train model weights whose distribution matches uniform and symmetric quantization functions.
Indeed, soft quantization methods do not have a gradient mismatch, but still omit fixed-point constraints. Furthermore, Bayesian training methods with quantizing priors do not reach the accuracy of deterministic soft quantization.

### *2.4. Summary of contribution*

In this paper we propose the SYMOG quantization approach, which does not require any retraining and can be used with pure fixed-point arithmetic. We claim the following contributions:

- Definition of a fixed-point constraint for symmetric and uniform quantization functions that replaces multiplications by bit shift operations (section 3.1).
- Introduction of a feasible loss function to train DNNs with multimodal weight distributions. The modes arrange themselves symmetrically around zero and minimize quantization error while taking into account the fixed-point constraint (section 3.2).
- Proposal of a new regularization parameter scheduling and a clipping function to enable self-reliant weight adaptation (section 3.3 and 3.4).
- Demonstration that our approach outperforms current 2-bit quantization approaches on three common benchmark data sets and varying model sizes (section 4.1 to 4.3).

## 3. Multimodal Fixed-point Weights

### *3.1. Fixed-Point Quantization*

The symmetrical representation of an $N$-bit fixed-point number is $(-1)^s \times m \times 2^{-f}$, where $(-1)^s \times m$ is the signed $N$-bit mantissa, $f \in \mathbb{Z}$ the position of the decimal point and $2^{-f}$ the global scaling factor (Sze et al. (2017)). Although the amount of feasible numbers is decreased by one, the symmetry property reduces DNN complexity on adequate hardware, see Faraone et al. (2018) and Krishnamoorthi (2018). Furthermore, in fixed-point arithmetic, a multiplication by a power of two is equivalent to moving the decimal point, respectively. This process, called bit shift, significantly accelerates computations on fixed-point hardware (Hennessy and Patterson (2017); Sze et al. (2017); Harris (2016)).

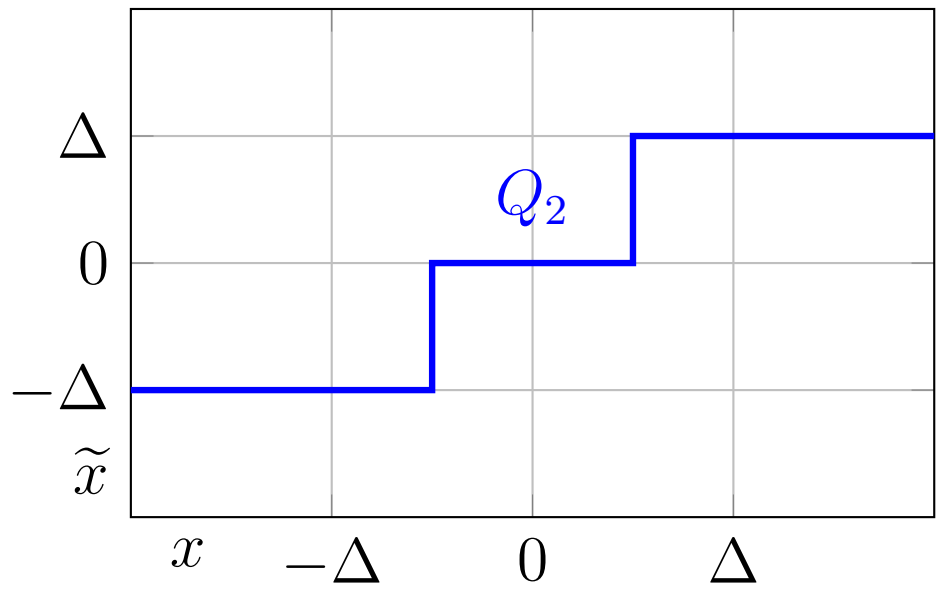

Figure 2: A symmetric and uniform 2-bit quantizer where $\Delta$ is the step size and $\widetilde{x}$ the quantized value of $x$.

On the other hand, a quantization function maps input signals to a smaller set of discrete values. If certain properties apply, the quantization function can be expressed with basic operations like scaling, rounding and clipping. For instance, a symmetric and uniform $N$-bit quantizer is

$$\widetilde{x} = Q_N(x; \Delta) = \underbrace{\mathrm{Clip}\left(\left\lfloor \frac{x}{\Delta} \right\rceil, -2^{N-1}+1, 2^{N-1}-1\right)}_{N\text{-bit signed integer}} \Delta\,, \tag{1}$$

where $\lfloor \cdot \rceil$ describes the rounding to the closest integer, $\Delta$ is the uniform step size, Clip(x, min, max) truncates all values to the domain [min, max] and $\widetilde{x}$ represents the quantized value. As an example, Figure 2 illustrates the symmetric 2-bit quantizer $Q_2$.

Obviously, Equation 1 corresponds to the fixed-point representation if and only if the step size is a power of two, hence $\Delta = 2^{-f}$, $f \in \mathbb{Z}$. In that case, $\widetilde{x}$ can be processed as a fixed-point number and the bit shift replaces the scaling operation. SYMOG uses $Q_N$, as defined in Equation 1, with $\Delta = 2^{-f}$ and $f \in \mathbb{Z}$ as a fixed-point constraint.

*3.2. Multimodal Gaussian Prior*

A DNN parameterizes a computational graph as a set of model parameters $\Theta = \{w_1, \ldots, w_M\}$ with $M$ being the number of parameters. In supervised learning, the DNN maps some input $X$ to an output variable $\widehat{Y}$ to approximate a set of known outcomes $Y$ (Goodfellow et al. (2016)). In order to find a suitable mapping, gradient descent progressively optimizes $\Theta$ along a cost function $C(Y, \widehat{Y})$, scaled by the learning rate $\eta$. An additional regularization term $R(\Theta)$ affects the probability distribution of $\Theta$ by adding a second gradient to the optimization step. Thus, the parameter update represents a trade-off between the learning task and the regularization and is controlled by the regularization parameter $\lambda$:

$$w \leftarrow w - \eta \left( \frac{\partial C}{\partial w} + \lambda \frac{\partial R}{\partial w} \right). \tag{2}$$

$L^2$-norm regularization is often used to prevent high filter energies and to give a Gaussian prior on the network weights (Krogh and Hertz (1991)). In order to combine multiple Gaussian modes and the fixed-point quantizer, we propose the following regularization term:

$$R = \sum_{l=1}^{L} \sum_{i=1}^{M_l} \frac{1}{M_l} \left\| w_{l,i} - Q_N\left(w_{l,i}; \Delta_l\right) \right\|_2^2 = \sum_{l=1}^{L} \sum_{i=1}^{M_l} \frac{1}{M_l} \left(w_{l,i} - \widetilde{w}_{l,i}\right)^2 , \tag{3}$$

where $L$ is the number of layers, $M_l$ the number of weights in layer $l$, $\widetilde{w}_{l,i}$ the quantized version of $w_{l,i}$, and $\Delta_l = 2^{-f_l}, f_l \in \mathbb{Z}$, the layer-wise fixed-point constraint.

Effectively, $R$ gives individual Gaussian priors to each network weight. The priors are updated with every forward pass with respect to the closest fixed-point cluster, enabling the weights to continuously switch between neighboring modes. The gradient of $R$ with respect to $w_{l,i}$ is as follows

$$\frac{\partial R}{\partial w_{l,i}} = \frac{2}{M_l} \left(w_{l,i} - \widetilde{w}_{l,i}\right) \underbrace{\left(1 - \frac{\partial \widetilde{w}_{l,i}}{\partial w_{l,i}}\right)}_{=1} = \frac{2}{M_l} \left(w_{l,i} - \widetilde{w}_{l,i}\right) . \tag{4}$$

Due to real-valued weights and a unique quantization function, the partial derivative of $Q_N$ can be assumed to be zero, see Figure 2. The final gradient is a scaled version of the corresponding quantization error. This property is beneficial since $Q_N$ does not have to be smooth and can, nevertheless, ensure that a fixed-point representation is achieved.

### 3.3. Regularization parameter

According to Equation 2, the regularization parameter $\lambda$ controls the weighting between the learning task on one hand and the fixed-point prior on the other. The larger $\lambda$ is, the stronger the weights are pushed in the direction of the closest fixed-point mode. To increase the model capacity, we start with a small value and increase $\lambda$ during the training. For this, we recommend exponential growth as described in Algorithm 1.

### 3.4. Weight clipping

Soft quantization approaches apply quantization constraints during the training to promote posterior distributions that are well qualified for post quantization. In SYMOG, the fixed-point constraint from Equation 1 limits the potential solution set of the layer weights to the discrete values $\pm\Delta_l\big(2^{N-1}-1\big)$. Obviously, weights should not exceed this interval during training. For example, a setup with $N=2$ bits and $\Delta_l=1$ leads to the quantization values $\{-1,0,1\}$. Once a weight value has reached $-1$, it is useless to optimize it in the negative direction because it would veer away from the solution set. The same concept applies on the opposite side. Therefore, we clip all weights within $\big[-\Delta_l\left(2^{N-1}-1\right), \Delta_l\big(2^{N-1}-1\big)\big]$ after each update step to promote weight adaptations that are constructive in terms of the fixed-point constraint. In section 4 we will demonstrate the benefit of the weight clipping in more detail.

### 3.5. Implementation

Algorithm 1 summarizes our SYMOG model to train DNNs with symmetric Gaussian priors that enable an accurate fixed-point quantization. A pretrained model as well as a learning-rate domain are required as input. Furthermore, a start value and a growth-factor must be specified for the regularization parameter. Considering a training time of $E$ epochs, we recommend $[\eta_0,\eta_E]=[0.01,0.001]$, $\lambda_0=10$, and $\alpha_E=9/E$.

## 4. Experiments

We use a setup with $N=2$ bits, which is a corner case since layer weights consist of ternary-coded vectors from $\{-1,0,1\}$ and the uniform step-size $\Delta$. An example of such a distribution is shown in Figure 1 on the right. Ternary-coded vectors replace multiply-accumulate operations by

**Algorithm 1** SYMOG: Training an $L$-layers DNN with fixed-point weights

**Input**: Pretrained model $M_\Theta$, Number of Epochs $E$, Training Data $(X, Y)$, Batch size $S$, Cost function $C$, Learning-rate domain $[\eta_0, \eta_E]$, Regularization start $\lambda_0$ and growth-factor $\alpha_E$, Desired bit-size $N$.

**for** $l = 1$ to $L$ **do**
  min. $||W_l - Q_N(W_l, \Delta_l)||^2$ // Calc. step size
  s.t. $\Delta_l = 2^{-f}, f \in Z$
**end for**
**for** $e = 1$ to $E$ **do**
  $\eta \leftarrow \eta_0 - (\eta_0 - \eta_E)\, e/E$ // Learning-rate
  $\lambda \leftarrow \lambda_0 \cdot \exp(\alpha_E\, e)$ // Reg. param
  Randomly shuffle $(X, Y)$ and
  partition into $B$ batches of size $S$
  **for** $b = 1$ to $B$ **do**
    $\widehat{Y}_b \leftarrow M_\Theta(X_b)$ // Forward pass
    $\frac{\partial C}{\partial w} \leftarrow \text{Backpropagate}\Big(\frac{\partial C(Y_b, \widehat{Y}_b)}{\partial \widehat{Y}_b}\Big)$ // Backward pass
    **for** $l = 1$ to $L$ **do**
      $\frac{\partial R}{\partial W_l} \leftarrow \frac{2}{M_l}\left(W_l - Q_N(W_l, \Delta_l)\right)$ // Quant. gradient
      $W_l \leftarrow W_l - \eta\left(\frac{\partial C}{\partial W_l} + \lambda \frac{\partial R}{\partial W_l}\right)$ // Update step
      $W_l \leftarrow \text{Clip}\big(W_l, \mp\Delta_l(2^{N-1} - 1)\big)$ // Weight clipping
    **end for**
  **end for**
**end for**
**for** $l = 1$ to $L$ **do**
  $\widetilde{W_l} \leftarrow Q_N(W_l, \Delta_l)$ // Quant. weights
**end for**
**return** $\{\widetilde{W_l}\}_{l=1}^{L}$

| Data set | Method | Model | Param. | Bits | Fixed-Point | Epochs | Error |
|---|---|---|---|---|---|---|---|
| **MNIST** | BC, Courbariaux et al. (2015) | - | - | 1 | yes | - | 1.29% |
| | TWN, Lin et al. (2016) | LeNet5 | 60k | 2 | no | 40 | 0.65% |
| | VNQ, Achterhold et al. (2018) | LeNet5 | 60k | 2 | no | 195 | 0.73% |
| | **SYMOG** | **LeNet5** | **60k** | **2** | **yes** | **25** | **0.63**% |
| | Baseline | LeNet5 | 60k | 32 | no | 25 | 0.7% |
| **CIFAR-10** | BC, Courbariaux et al. (2015) | VGG8 | 14M | 1 | yes | 500 | 9.90% |
| | TWN,Lin et al. (2016) | VGG7 | 12M | 2 | no | 170 | 7.44% |
| | **SYMOG** | **VGG7** | **12M** | **2** | **yes** | **100** | **5.71**% |
| | Baseline | VGG7 | 12M | 32 | no | 100 | 5.52% |
| | VNQ, Achterhold et al. (2018) | DenseNet | 0.49M | 2 | no | 150 | 8.83% |
| | **SYMOG** | **DenseNet** | **0.49M** | **2** | **yes** | **100** | **5.96**% |
| | Baseline | DenseNet | 0.49M | 32 | no | 100 | 5.72% |
| **CIFAR-100** | TWN,Lin et al. (2016) | VGG11 | 32M | 2 | no | 300 | 36.18% |
| | BR, Yin et al. (2018) | VGG11 | 32M | 2 | no | 300 | 34.13% |
| | **SYMOG** | **VGG11** | **32M** | **2** | **yes** | **100** | **32.05**% |
| | Baseline | VGG11 | 32M | 32 | no | 100 | 31.42% |
| | TWN,Lin et al. (2016) | VGG16 | 34M | 2 | no | 300 | 28.59% |
| | BR, Yin et al. (2018) | VGG16 | 34M | 2 | no | 300 | 27.90% |
| | **SYMOG** | **VGG16** | **34M** | **2** | **yes** | **100** | **27.65**% |
| | Baseline | VGG11 | 34M | 32 | no | 100 | 26.58% |

Table 1: Summary of the quantized 1-bit (BC) and 2-bit (TWN, VNQ, BR, SYMOG) performance on MNIST, CIFAR-10 and CIFAR-100. The baseline gives the 32-bit floating-point accuracy of the particular model.

additions and subtractions (Courbariaux et al. (2015)) whereas the multiplication with $\Delta$ results in a bit shift operation. Therefore, multiplications can be prevented, which significantly accelerates the layer computation. The experiments are organized in two parts:

1. The performance of SYMOG is evaluated on three popular benchmark data sets (MNIST, CIFAR-10, CIFAR-100) using different network architectures from small to large-scale (section 4.1 to 4.3).
2. We give insights in different stages of SYMOG training and illustrate the self-reliant weight adaptation (section 4.4).

All experiments are performed using Algorithm 1 and stochastic gradient descent optimization with Nesterov momentum 0.9 and a batch size of 64. We compare SYMOG with weight compression results from BinnaryConnect (BC, Courbariaux et al. (2015)), Ternary Weight Networks (TWN, Li and Liu (2016)), Variational Network Quantization (VNQ, Achterhold et al. (2018)), and BinaryRelax (BR, Yin et al. (2018)). The results are shown in Table 1. As usual, we initialize the network weights with an accurate floating-point model which also provides the 32-bit floating-point baseline in Table 1.

### *4.1. LeNet-5 on MNIST*

MNIST is a handwritten-digits classification task with 28×28 gray scale images, divided into 50,000 training and 10,000 test samples (LeCun and Cortes (2010)). We preprocess the data by subtracting the mean and dividing by the standard-deviation over the training set. We use LeNet-5 from Lecun et al. (1998) and train for 25 epochs. Consequently, we achieve the lowest test error of 0.63% and even outperform the 0.7% baseline. This proves SYMOG's characteristic as a regularization method. The amount of training epochs is clearly higher for TWN and VNQ with 40 and 195 epochs, respectively.

### *4.2. VGG7 & DenseNet on CIFAR-10*

CIFAR-10 is an image classification benchmark data set, consisting of 32×32 colored pictures with 50,000 training and 10,000 test samples (Krizhevsky (2009)). We follow the same preprocessing steps as in Huang et al. (2016) and test two different network architectures. First, a conventional CNN with seven layers and batch normalization called VGG7 (Simonyan and Zisserman (2014)). Second, a modern DenseNet (L = 76, k = 12, Huang et al. (2016)) which has an optimized architecture with comparatively less parameters. Due to its lower number of redundancies, DenseNet is described as difficult to quantize (Achterhold et al. (2018)).
Our VGG7 achieves the lowest values in both test error and training epochs. We achieve 5.71% test error, which outperforms TWN by more than 1.5% and is almost equal to the 32-bit baseline of 5.52%. The training effort of 100 epochs is clearly lower compared to 170 epochs of TWN training.
With the DenseNet architecture, SYMOG achieves 5.96% test error after 100 training epochs and clearly outperforms the Bayesian approach of VNQ-DenseNet by nearly 3%. The 5.72% high-precision baseline is close to our result which proves that our approach works well for small architectures.

### *4.3. VGG11 & VGG16 on CIFAR-100*

CIFAR-100 uses the same images as CIFAR-10 but provides 10 additional sub-classes for each class in CIFAR-10. Thus, only 500 training samples and 100 test samples are available for each of the 100 classes, which makes CIFAR-100 a challenging classification task. We use the same preprocessing steps as in Huang et al. (2016) in combination with VGG11 and VGG16 from Simonyan and Zisserman (2014). We compare with the latest results from TWN and BR.
After 100 training epochs, our VGG11 reaches 32.05% test error and outperforms BR-VGG11 by more than 2% and TWN-VGG11 by more than 4% test error, while both BN and TWN train significantly longer, i.e. 300 epochs. Thus, SYMOG improves the quantized training of DNNs in both effort and accuracy.
Considering the VGG16 experiment, SYMOG performs best using symmetric ternary-valued weights as well. The test error is 27.65%, which is 0.25% less than BR-VGG16 at one third of the training time.
Consequently, SYMOG is able to reduce the quantized training for CIFAR-100 by a factor of three while improving the accuracy by up to 4%.

### *4.4. Weight adaptation*

SYMOG performs well on the soft quantization task of DNNs. Even in the case of 2-bit fixed-point weights, SYMOG is able to maintain most of the 32-bit accuracy in constant training time. In this section, we give detailed insights into the training process of VGG11 on CIFAR-100. We analyze the three-modal Gaussian prior and the self-reliant weight adaptation.
As an example, Figure 3 shows the weight distribution of Layer-1, Layer-4 and Layer-7 after several epochs of training. Since weight decay is used for pretraining, the initial weights resemble a unimodal distribution with a single peak at zero (epoch 0). At the start of training, two additional peaks arise at $\pm\Delta$ since layer weights are clipped to the particular quantization domain. Following this, the weights start to rearrange. The adaptation process in single layers can be seen in Figure 4 (upper plot), which gives the percentage of weights that switch to another fixed-point prior during single epochs. Due to a small but exponentially increasing regularization parameter, the Gaussian priors are restricted at the beginning, favoring and increased fluctuation rate. On average, 22% of the weights in Layer-7 change their fixed-point annotation during each epoch in the first half of training. After 80 epochs, three separated Gaussian modes are clearly visible, but 1.8% of the weights still change their fixed-point cluster by the end of training.
However, weight adaptation behaves differently from layer to layer. For example, Layer-1 is completed first after 32 epochs, whereas weights in Layer-10 are adapting until the end of the training. This may come from different step sizes and layer-dependent gradient scales.
In order to analyze the influence of the weight clipping described in Section 3.4, the bottom plot in Figure 4 shows the weight adaptation if the clipping is skipped. Initially, one can see that the absolute number of changing modes is clearly lower. Thus, the average changing rate of Layer-7 is around 8% in the first half of training compared to 22% with clipping. This holds also for other layers or networks we have tested. Furthermore, one can observe that the changing rate of Layer-4, Layer-5 and Layer-7 increases once again after 40 epochs. This is due to weights that lie out of the quantization domain and must first pass the distance to the outlying modes at $\pm\Delta$.

## 5. Conclusion & Outlook

We propose SYMOG, a Gaussian prior for multiple fixed-point modes and we address two common problems in weight quantization: high-precision scaling coefficients and retraining. The fixed-point constraint for symmetric and uniform quantization functions replaces multiplications by bit shift operations. With exponentially increasing Gaussian priors and weight clipping, SYMOG provides self-reliant weight adaptation. In multiple experiments we demonstrate the benefit of SYMOG in terms of training epochs and accuracy by achieving new state-of-the-art results on MNIST, CIFAR-10 and CIFAR-100. We provide insights into the training procedure and illustrate the effect of single components.
For the future, we will investigate the individual adaptability of single lay-ers in more detail. Further, we will extend the gradient-based fixed-point training to remaining layer-types such as batch normalization, dropout and ReLU. In this way, we can finally generate pure fixed-point models.

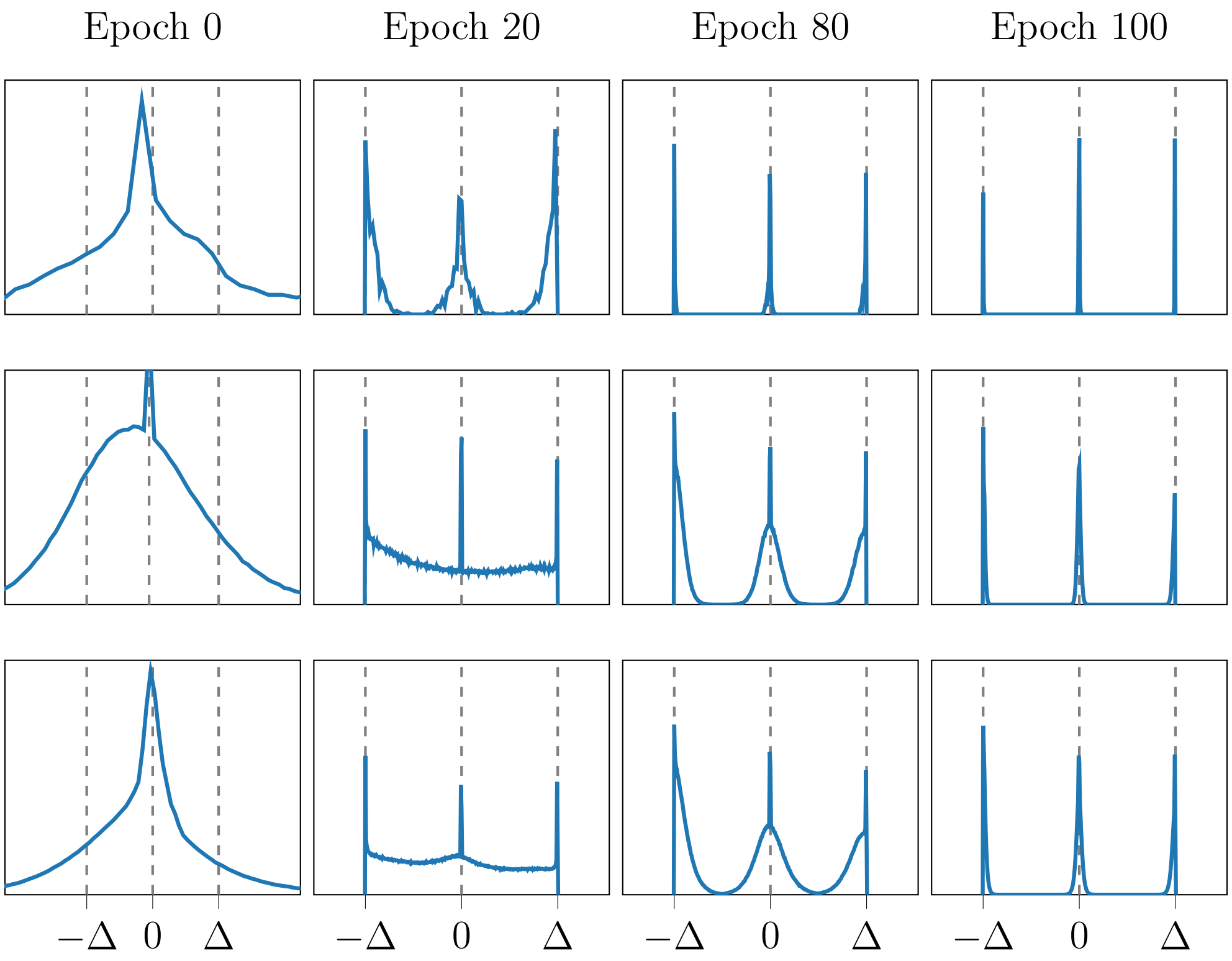

Figure 3: Weight distribution of Layer-1, Layer-4 and Layer-7 (from top to bottom) of VGG11 after several epochs. Since weight decay is used for pre-training, the weight distribution is unimodal at the beginning with a peak at zero. Then, SYMOG continuously rearranges the weights into a three-modal Gaussian distribution, clearly visible at epoch 80. The variance of each mode is continuously decreased by the exponentially increasing regularization parameter. After 100 epochs, the weights are that close to the fixed-point centers that post quantization does not produce a remarkable error. Note: y-axis scaled individually for convenience, the x-axis for epoch 0 is wider to catch the whole distribution.

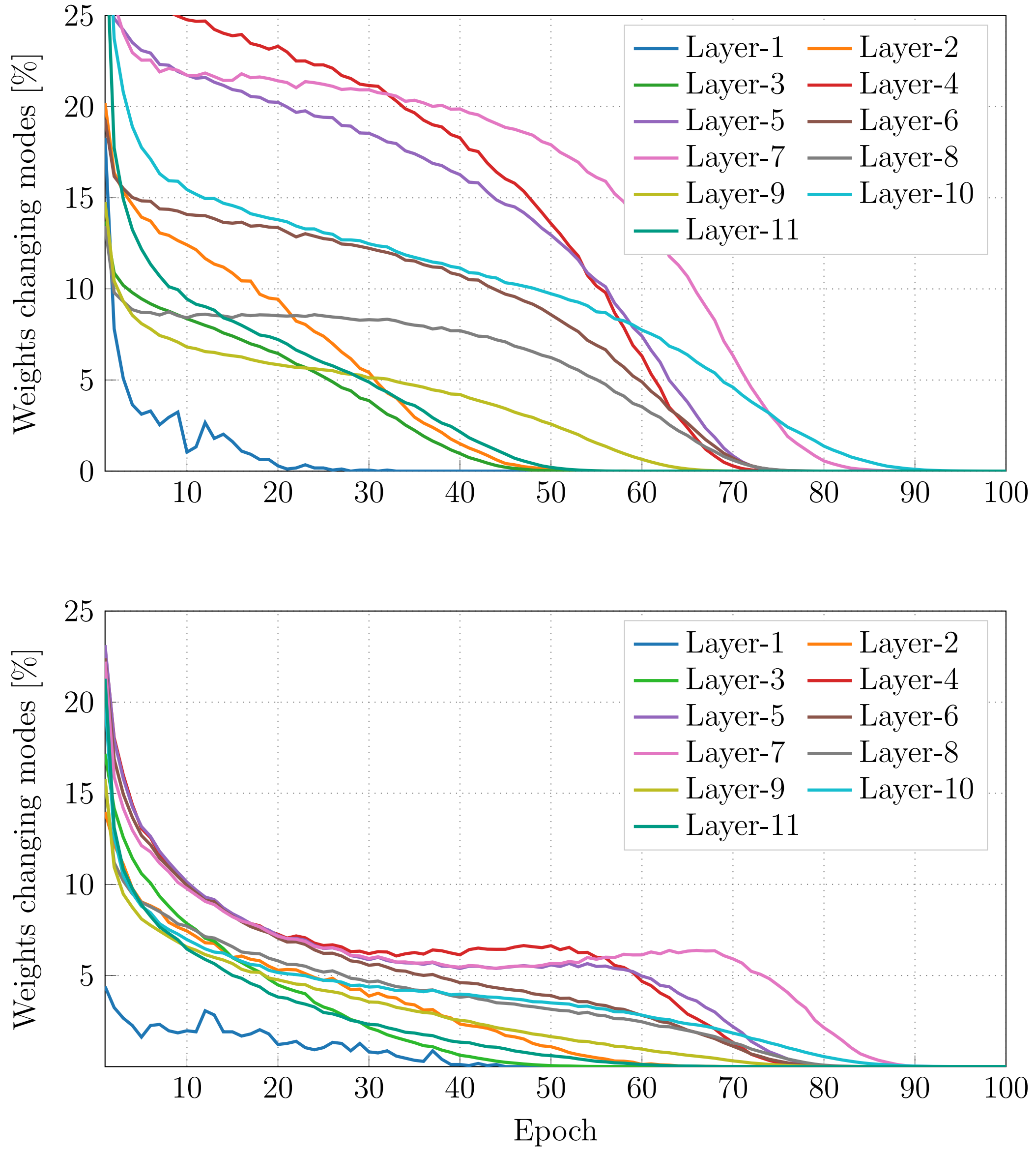

Figure 4: Both plots illustrate the weight adaptation in SYMOG training. The y-axis gives the percentage of weights that change their fixed-point prior (which is the closest fixed-point cluster) during a single epoch. The upper plot results if weight clipping is used as described in Algorithm 1, the lower plot results if the clipping is skipped. One can observe that the weight adaptation is clearly improved by the clipping. Especially in the very beginning of training, many weights rearrange. Thus, the weight clipping improves SYMOG in both accuracy and training time.